\documentclass[runningheads]{llncs}

\usepackage[T1]{fontenc}
\usepackage{graphicx}
\usepackage{amsmath}
\usepackage{amssymb}
\usepackage{booktabs}
\usepackage[table]{xcolor}
\usepackage{multirow}
\usepackage{enumitem}
\usepackage[hidelinks]{hyperref}

\setlist{nosep,leftmargin=1.2em}
\begin{document}

\title{Auto-Director: Leveraging Multi-Agent Collaboration for
Autonomous Video Editing}


\author{Yuxia Chen\inst{1} \and
Wenbo Zhu\inst{2} \and
Yangguang Ji\inst{3} \and
Zhenxiang Jiang\inst{4} \and
Teng Zhong\inst{5} \and
Yu Zhang\inst{6}\thanks{Corresponding author.}}
\authorrunning{Y. Chen et al.}

\institute{University of Electronic Science and Technology of China, China \and
University of California, Berkeley \and
Nanyang Technological University \and
National University of Singapore \and
Nanjing Normal University \and
Beijing Freedo Technology}

\titlerunning{Auto-Director: Multi-Agent Collaboration for Autonomous Video Editing}

\maketitle

\begin{abstract}
Autonomous video editing from open-world resources remains a challenging problem in multimedia content
creation. Retrieval-based methods are constrained by fixed source material and lack creative flexibility,
while generative models struggle with long-horizon coherence and often diverge from user intent. Both paradigms fail to replicate the structured, multi-stage reasoning that underlies professional post-production. To address these limitations, we present Auto-Director, a tool-augmented multi-agent framework that
mirrors professional editorial workflows: a Planning Agent decomposes abstract topics into fact-grounded
scripts via retrieval-augmented planning; a Material Agent indexes open-world footage through shot
segmentation and visual tagging; an Editor Agent assembles semantically aligned clips with MLLM-driven
spatial adaptation; and a Critic Agent governs an iterative Reflect-and-Revise loop that checks scene-level
alignment and pacing. To rigorously evaluate this new paradigm, we introduce VEdit-Bench, a benchmark of 100 expert-curated
topics with human-annotated Golden Scripts and a five-dimensional metric suite covering Narrative Logic,
Factual Accuracy, Semantic Alignment, Compositional Quality, and Stylistic Cohesion. Extensive experiments demonstrate that Auto-Director significantly outperforms baseline methods in both automated metrics and human evaluation, effectively bridging the gap between amateurs and professional-grade video production.

\end{abstract}

\section{Introduction}
\label{sec:intro}
Video has become a dominant medium for information dissemination and storytelling, yet professional
production remains labor-intensive, demanding expertise in scriptwriting, material sourcing, and aesthetic
editing~\cite{wu2025video}, which places narrative-driven production out of reach for most amateurs.

\begin{figure}[t]
 \centering
 \includegraphics[width=0.82\linewidth]{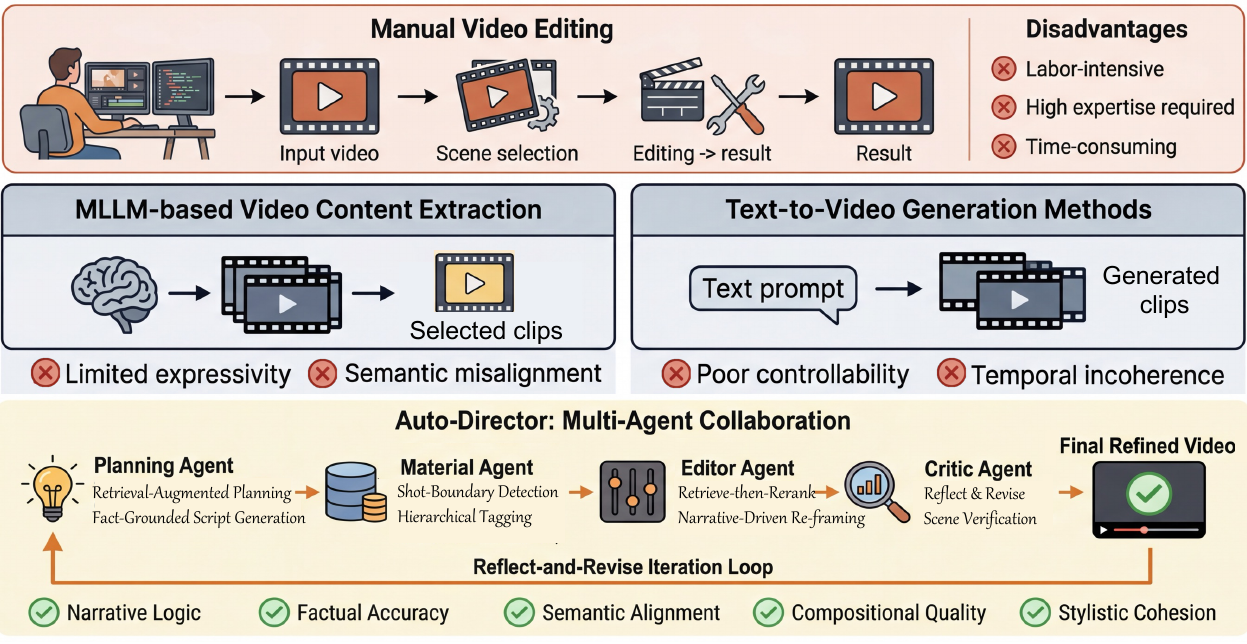}
 \caption{Overview of Auto-Director: a structured multi-agent pipeline with iterative reflect-and-revise
  feedback, addressing the limited expressivity of retrieval-based and the weak controllability of
  generation-based paradigms.}
 \label{fig:teaser}
\end{figure}

Recent advances in AI have fostered two primary directions for automated video creation, and both face
limitations for open-world, long-form storytelling (see Fig.~\ref{fig:teaser}). Retrieval-based methods
compose highlights from a fixed source corpus, as in trailer generation~\cite{argaw2024towards}; grounded in
real footage, they are constrained by their source material and cannot easily bridge the semantic gap between
nuanced script requirements and coarse dataset metadata. Generation-based methods instead synthesize clips
directly from text~\cite{bar2024lumiere,wu2025unlearning}; although increasingly capable on visual quality, cinematic
multi-shot generation still treats structural control and cross-shot consistency as open
challenges~\cite{zhang2025stage,wu2026kris}, and such methods remain vulnerable to long-horizon inconsistency, factual
hallucination, and weak controllability over many scenes. A third, more recent line applies agentic decomposition directly to video
editing~\cite{li2026direct,lin2026glance,zhou2026videoagent}.
These systems emphasize different parts of the workflow; none combines externally grounded factual planning,
structured shot-level indexing, and an iterative quality-control loop in one pipeline.

To address these limitations, we present Auto-Director, an agentic framework for open-world autonomous video editing that emulates the
professional post-production workflow through four specialized agents. In contrast to human-in-the-loop
language-assisted editors such as LAVE~\cite{wang2024lave}, our goal is full pipeline autonomy. A Planning
Agent translates abstract topics into fact-grounded scripts via Retrieval-Augmented Planning, decoupling
factual accuracy from language fluency. A Material Agent converts open-world footage into structured,
searchable assets through shot boundary detection and explicit visual tagging, reducing cross-modal retrieval
to text-to-text matching. An Editor Agent performs two-stage semantic matching and MLLM-driven spatial
adaptation to synthesize a coherent draft. Finally, a Critic Agent then governs a Reflect-and-Revise loop, dispatching
structured feedback upstream to check alignment and pacing.

To evaluate this setting, we introduce VEdit-Bench, a benchmark of 100 expert-curated topics paired with
human-annotated Golden Scripts and fragmented material repositories that require open-world retrieval, with a
five-dimensional metric suite. Our contributions are summarized as follows:
\begin{itemize}[leftmargin=1em,topsep=0pt]
\setlength{\itemsep}{0pt}
\item We propose Auto-Director, a fully automated multi-agent framework orchestrating four specialized agents
to produce narrative videos from open-world resources, with an iterative Reflect-and-Revise mechanism that
self-corrects across retrieval, composition, and pacing.
\item We construct VEdit-Bench, a benchmark of 100 expert-curated topics with human-annotated Golden Scripts
and fragmented material repositories, paired with a five-dimensional metric suite including Factual Accuracy,
a dimension prior video generation benchmarks do not explicitly target.
\item Extensive experiments demonstrate that Auto-Director significantly outperforms leading baselines across all five metrics, effectively bridging the gap between amateur creators and professional-grade video production.
\end{itemize}

\section{Related Work}
\label{sec:related_work}

\subsection{Narrative Video Creation}

Progress on narrative video creation spans controllable generation and automated editing. Story visualization exposed the need for global narrative consistency beyond isolated frame synthesis, and
recent methods target explicitly cinematic multi-shot generation: STAGE~\cite{zhang2025stage,wu2025number} introduces
storyboard-anchored generation with cross-shot memory, and related work studies long-form storytelling
through adaptive memory and staged generation. These remain fully generative and therefore cannot ground a
scene in verifiable real-world evidence.

A complementary line edits or composes existing footage: TGT~\cite{argaw2024towards} formulates trailer
creation as sequence-to-sequence shot prediction over a single movie,
AesopAgent~\cite{wang2024aesopagent} organizes story-to-video generation as multi-agent orchestration, and
LAVE~\cite{wang2024lave} shows how LLMs can assist human editing. These improve controllability and workflow
automation, but either remain fully generative, assume closed-world assets, or keep the human editor in the
loop at execution time.

\subsection{Agentic Content Creation}


Agentic decomposition has proven effective for multimodal content creation more broadly, typically separating
planning, rendering, and self-critique roles. World-To-Image~\cite{son2025world,wu2024glance,wu2026generalization}, for example, grounds
text-to-image generation in agent-driven web retrieval. Similar agentic designs have recently been explored
for video editing, where planning, material organization, editing, and refinement need to be coordinated.

EditDuet~\cite{sandovalcastaneda2025editduet} pairs an Editor with a Critic that provides natural-language
feedback over input clips, while a prompt-driven agentic editor~\cite{ding2025promptdriven} builds a global
narrative index for long-form media through temporal segmentation and memory compression.
DIRECT~\cite{li2026direct} adopts hierarchical multi-agent planning for video mashup creation, and
GLANCE~\cite{lin2026glance} coordinates global and local loops for music-grounded editing.
CineAgents~\cite{zhang2026cinebench} studies instruction-driven cinematic compilation with an expert-annotated
benchmark, while VideoAgent~\cite{zhou2026videoagent} uses intent parsing and graph-based tool orchestration
for general-purpose video editing. Auto-Director follows this agentic direction by integrating factual
planning, structured material processing, editing, and iterative refinement within a unified narrative video
editing workflow.

\section{Auto-Director}
\label{sec:method}

We formalize the Auto-Director pipeline as a structured transformation from a user prompt to a finished
narrative video. The overall framework is illustrated in Fig.~\ref{fig:teaser}.

\subsection{Planning Agent}
\label{sec:planning}

The Planning Agent serves as the directorial brain of Auto-Director, translating an abstract user prompt $P$ into a structured, executable script $S$. The core
challenge in this stage is factual grounding, since LLMs hallucinate on knowledge-intensive topics where precise attribution
is non-negotiable~\cite{ji2023towards,zhu2026lookcarefully,zhu2026mitigating}. Rather than relying
on the model's parametric knowledge alone, we integrate a
Retrieval-Augmented Planning mechanism inspired by retrieval-augmented
generation~\cite{lewis2020retrieval}: the agent decomposes $P$ into atomic sub-queries $\{q_k\}$, issues each to
external knowledge APIs, and retains only claims corroborated by the retrieved evidence, giving a verified fact
set $\mathcal{K}$ on which all generation is conditioned; the agent may not introduce dates, names, or causal
relations unsupported by it. This grounding step decouples factual accuracy from language
fluency, ensuring the script is both coherent and verifiable.

The agent transforms $\mathcal{K}$ into a script $S = \{s_i\}_{i=1}^{N}$ under a fixed JSON schema, so
downstream agents parse it deterministically, with each scene a three-tuple $s_i = (T_i, Q_i, D_i)$:
\begin{itemize}[leftmargin=1em,topsep=0pt]
\setlength{\itemsep}{0pt}
 \item \textbf{Fact-Anchored Narration} $T_i$: voiceover text synthesized from $\mathcal{K}$.
 \item \textbf{Visual Query Description} $Q_i$: a caption specifying the desired clip content for downstream
 retrieval; abstract ideas are converted into named events, persons, objects, or places to make them
 searchable.
 \item \textbf{Temporal Constraint} $D_i$: a duration window governing scene pacing.
\end{itemize}

Notably, the Planning Agent participates not only in the initial generation
phase but also in the Critic Agent's refinement loop. When no suitable footage
exists for a given scene $s_i$, the Critic dispatches an adaptive scripting
signal $\mathcal{F}_{\text{script}}$ that triggers the Planning Agent to
revise the narration $T_i \rightarrow T'_i$ and query $Q_i \rightarrow Q'_i$, realigning the script with
available visual evidence rather than forcing an irreconcilable
visual-text mismatch.

\subsection{Material Agent}
\label{sec:grounding}

The Material Agent bridges abstract script requirements and raw open-world footage. Its design addresses two
limitations of existing retrieval paradigms: dual-encoder VLMs such as CLIP~\cite{radford2021learning,zhu2024enhancing,zhu2024selective,zhu2025enhancing} may lack
the world knowledge needed for knowledge-intensive concepts such as historical events or named entities
~\cite{tschannen2025siglip,zhu2024boosting,zhu2025dynamic,zhu2026hierarchical}, while MLLMs often localize event boundaries imprecisely
~\cite{huang2024vtimellm}. We therefore convert footage into a structured, text-indexed asset library
$\mathcal{D}$ before semantic matching.

\paragraph{\textbf{Step 1: Shot Segmentation and Visual Tagging}}
Each video is segmented into cinematically atomic shots via content-aware boundary detection
~\cite{pyscenedetect,soucek2024transnet}, yielding $V_{\text{shots}}=\{v_1,\ldots,v_k\}$. For each shot $v_j$, a tagged
frame is constructed by superimposing a structured text annotation directly
onto sampled keyframes:
\begin{equation}
    \tilde{f}_j = \operatorname{Overlay}(f_j,\; \text{tag}_j), \quad
    \text{tag}_j = [\,\texttt{Shot\#}j,\;\; t_j^{\text{time}}\,],
\end{equation}
where the overlay is rendered in a fixed format to ensure consistent
readability across diverse footage styles. This explicit visual tagging
embeds both structural and temporal information directly into the visual
domain~\cite{wu2024visual,chen2024hi,fang2026chasm}, enabling MLLMs to reliably read shot identifiers and timestamps
without relying on external metadata, which is a known source of
hallucination~\cite{wu2025number}.

\paragraph{\textbf{Step 2: Hierarchical Caption Generation}}
An MLLM processes the tagged keyframes and generates a structured JSON caption $d_j$ for each shot, including
\textbf{high-level context}, \textbf{shot-level details} such as entities, actions, and camera movements, and
\textbf{exact timestamps} read from the visual tags. The resulting captions form the asset library
$\mathcal{D}=\{d_j\}_{j=1}^{M}$.

\paragraph{\textbf{Step 3: Text-to-Text Semantic Indexing}}
With all captions aggregated, the asset library $\mathcal{D} = \{d_j\}_{j=1}^{M}$
is indexed by embedding each structured caption $d_j$ into a dense vector
representation via a text encoder, constructing a searchable index over
the textual domain. Downstream agents query $\mathcal{D}$ using natural
language descriptions rather than raw visual features, reducing cross-modal
retrieval to a text-to-text matching problem, which moderately
reduces inference overhead on visual models and mitigates
hallucination risks arising from insufficient cross-modal alignment.

\subsection{Editor Agent}
\label{sec:editor}

The Editor Agent transforms heterogeneous clips into a cohesive draft video $V_{draft}$ that matches the
user-specified aspect ratio. It operates in three sequential phases: two-stage semantic matching, MLLM-driven
spatial adaptation, and narrative-conditioned layout composition.

\paragraph{\textbf{Step 1: Two-Stage Retrieve-then-Rerank}}
To select the optimal footage for each scene $s_i$ from the candidate pool,
we propose a two-stage Retrieve-then-Rerank pipeline, following the general
dense retrieval and neural reranking paradigm established in open-domain
retrieval~\cite{reimers2019sentence,karpukhin2020dense,nogueira2019passage}. 
We first encode the visual query $Q_i$ with a text encoder $\phi(\cdot)$
(Qwen3-Embedding~\cite{zhang2025qwen3,zhu2026robustifying}) and retrieve the top-$K$ candidates from the material library
$\mathcal{D}$ using cosine similarity:
\begin{equation}
    \mathcal{C}_i^K = \underset{d_j \in \mathcal{D}}{\operatorname{top-}K}\;
    \frac{\phi(Q_i)^\top \phi(d_j)}{|\phi(Q_i)|\cdot|\phi(d_j)|}.
\end{equation}
The MLLM then acts as a cross-modal re-ranker. For each candidate $c_k \in \mathcal{C}_i^K$, it jointly
considers the narration $T_i$, structured caption $d_k$, and sampled keyframes $\mathcal{F}_k$, and scores
the candidate on three dimensions: \emph{semantic fidelity}, \emph{atmospheric coherence}, and
\emph{temporal suitability}. The scores are aggregated as
\begin{equation}
    r_k = \lambda_1 r_k^{\text{sem}} + \lambda_2 r_k^{\text{atm}} +
    \lambda_3 r_k^{\text{tmp}}, \quad
    c_i^* = \arg\max_{c_k \in \mathcal{C}_i^K} r_k,
\end{equation}
where $\lambda_1 = 0.5$, $\lambda_2 = 0.3$, and $\lambda_3 = 0.2$, prioritizing semantic correctness.
The two-stage design achieves a coarse-to-fine retrieval hierarchy: the embedding stage efficiently prunes the search space from $M$ to $K \ll M$, while the MLLM re-ranking stage contributes the high-level narrative reasoning that embedding-only models cannot provide.

\paragraph{\textbf{Step 2: MLLM-Driven Spatial Adaptation}}
After selecting $c_i^*$, the Editor Agent adapts the source footage to the target aspect ratio
$R_{\text{target}}$. The MLLM analyzes each frame together with the narration $T_i$ and predicts normalized
bounding boxes for the primary subject. To avoid unstable frame-wise cropping, we apply Kalman
filter-based stabilization~\cite{kalman1960new}, modeling ROI center coordinates as a linear dynamical system
to suppress high-frequency jitter while preserving smooth subject tracking
~\cite{cao2024reframe,chen2026hifi,shen2024prune}.

\paragraph{\textbf{Step 3: Narrative-Conditioned Layout Composition}}
The agent further determines whether a single crop can faithfully represent the scene. If the narration
$T_i$ jointly references $n \geq 2$ spatially disjoint subjects whose bounding boxes have insufficient overlap,
a single crop may omit important visual information. The agent therefore selects a multi-window layout, such
as Split-Screen or Picture-in-Picture, assigning separate windows to the relevant subject regions. Otherwise,
the stabilized single-stream crop is retained.
\subsection{Critic Agent}
\label{sec:refinement}

Linear pipelines propagate errors, with a single retrieval failure or misaligned crop degrading the output and
no mechanism for recovery. Auto-Director addresses this with a Reflect-and-Revise loop governed by an MLLM-based
Critic Agent, which audits $V_{draft}$ against $S$ and dispatches structured feedback upstream, following
iterative self-feedback paradigms~\cite{madaan2023self,zhu2026guardalign,shinn2023reflexion,zhu2026principled}. Each scene is evaluated
independently: the Critic emits integer semantic and aesthetic scores $v_i^{\text{sem}}$, $v_i^{\text{aes}}$ on
a $1$--$5$ scale plus a signal naming the scene, failure mode, and corrective action, so revision is surgical
rather than global, and each failure routes to exactly one signal type.

\paragraph{\textbf{Dimension 1: Semantic and Factual Verification}}
The Critic samples keyframes from $V_{draft}$ and computes $v_i^{\text{sem}}$ by jointly evaluating $T_i$ and
$Q_i$, where 5 indicates the footage precisely depicts the narration's entities, events, and relations and 1
that it fails entirely. On a factual mismatch where the clip is wrong but a replacement exists in
$\mathcal{D}$, the Critic emits $\mathcal{F}_{\text{retrieval}}$ (\emph{hallucination detection}) to trigger
re-retrieval with a refined query $Q'_i$; when no suitable footage exists, it dispatches
$\mathcal{F}_{\text{script}}$ (\emph{adaptive scripting}) to the Planning Agent.

\paragraph{\textbf{Dimension 2: Aesthetic and Temporal Audit}}
The Critic then evaluates cinematic quality via $v_i^{\text{aes}}$ on the same scale: cluttered or truncated
compositions trigger $\mathcal{F}_{\text{layout}}$, prompting the Editor Agent to adjust the ROI trajectory or
layout template, while abrupt cuts or violations of $\{D_i\}$ trigger $\mathcal{F}_{\text{timing}}$, prompting
adjustment of shot durations or transitions.

A scene passes iff both thresholds are satisfied, and passing scenes are frozen and excluded from later
cycles. The termination condition is
\begin{equation}
 t^* = \min\!\left\{t \leq \tau \;:\;
 \forall i,\;\; v_i^{\text{sem}(t)} \geq \tau_{\text{sem}}
 \;\wedge\; v_i^{\text{aes}(t)} \geq \tau_{\text{aes}}\right\},
\end{equation}
so the loop runs until all per-scene criteria are met or the budget $\tau$ is exhausted.

\section{VEdit-Bench}
\label{sec:benchmark}
In this section, we introduce VEdit-Bench, a benchmark designed to evaluate the full lifecycle of autonomous video editing. It combines expert-curated topics, human-annotated Golden Scripts, and fragmented material repositories to evaluate systems under realistic long-form editing scenarios.

\subsection{Benchmark Construction}

\paragraph{\textbf{Topic Selection and Taxonomy}}
We manually curated 100 diverse topics from trending educational and documentary categories on global video
platforms and filtered them through a panel of five human experts. The topics follow a three-tiered difficulty
taxonomy and are distributed as 40 Easy / 40 Medium / 20 Hard:
(1) \emph{Entity-Centric}, centered on concrete visual entities (e.g., ``The Life Cycle of a Monarch
Butterfly'');
(2) \emph{Abstract Concept}, requiring cross-domain reasoning to translate non-visual ideas into coherent
imagery (e.g., ``The Impact of AI on Modern Society''); and
(3) \emph{Complex Narrative}, requiring strict temporal logic and identification of specific historical or
technical entities (e.g., ``The Rise and Fall of the Roman Empire'').

\paragraph{\textbf{Expert-Annotated Golden Scripts}}
For each of the 100 topics, professional scriptwriters composed a Golden Script ($S_{gold}$) containing precise narration, fact-anchored milestones, and required visual evidence descriptions. Each script was validated by three additional expert annotators (Fleiss' $\kappa = 0.74$), and serves as the ground-truth reference for video evaluation.

\paragraph{\textbf{Fragmented Material Repositories}}
For each topic, we provide local video assets that deliberately cover only a partial subset of the required
visual milestones, averaging 58\% coverage of those defined in $S_{gold}$. This fragmented setting reflects
practical editing scenarios in which the available footage may not fully satisfy the narrative requirements,
requiring the system to retrieve missing evidence from external sources.

\subsection{Evaluation Metrics}
We propose a five-dimensional metric suite. All metrics are evaluated by prompting an
MLLM~\cite{chen2024mllm}. The judge uses sampled keyframes with task-specific references, including $S_{gold}$ for Factual Accuracy
and generated narration for Semantic Alignment. All systems are evaluated with the same protocol, and the agreement between MLLM and human
judgments is further validated in Sect.~\ref{sec:user_study}.

\paragraph{\textbf{Narrative Logic}}
The MLLM processes sampled keyframes together with the generated script and evaluates narrative flow along
three dimensions: \emph{causal progression}, whether each scene follows logically from the previous one;
\emph{thematic consistency}, whether the overall topic remains focused without unnecessary digression; and
\emph{structural coherence}, whether the narrative forms a clear introduction, development, and conclusion.
Narrative Logic therefore captures high-level failures such as abrupt logical transitions, unresolved
narrative threads, and contradictions between adjacent scenes.

\paragraph{\textbf{Factual Accuracy}}
The MLLM cross-references sampled keyframes against the fact-anchored milestones and required visual evidence
in $S_{gold}$, verifying whether the depicted entities, objects, structures, and events are objectively
correct regardless of what the model-generated narration states. This metric targets visual factual errors
such as incorrect historical figures, species, artifacts, or visually similar but factually distinct
landmarks. It therefore evaluates correctness against the Golden Script rather than consistency with the
system's own narration.

\paragraph{\textbf{Semantic Alignment}}
For each scene $s_i$, a MLLM evaluates sampled keyframes alongside
the model-generated narration $T_i$, assessing whether the described entities, actions, and events are physically manifested in the video at the correct temporal position. Unlike Factual Accuracy, which measures correctness against the ground-truth $S_{gold}$, Semantic Alignment measures internal consistency between the model's own narration and the retrieved footage, and can thus diverge from Factual Accuracy: a scene may achieve high SA when the footage
faithfully supports an incorrect narration, or low SA when correct footage is paired with a misaligned description. The final SA score is averaged across all $N$ scenes.

\paragraph{\textbf{Compositional Quality}}
The MLLM scores each clip along three aspects: \emph{subject preservation}, whether the primary subject
described in $T_i$ remains correctly framed within the target aspect ratio; \emph{spatial balance}, whether
the composition avoids perceptual imbalance or excessive dead space; and \emph{cropping artifact avoidance},
whether re-framing unintentionally removes salient visual regions. Compositional Quality therefore reflects
the within-shot layout quality produced by the spatial adaptation stage, with scores averaged over all
$N$ scenes.

\paragraph{\textbf{Stylistic Cohesion}}
The MLLM assesses the global visual consistency of the assembled video by examining whether the dominant
visual style, such as cinematic, archival, or animated footage, remains coherent across clips. Unlike
Compositional Quality, which operates at the individual-shot level, Stylistic Cohesion is a sequence-level
metric that penalizes abrupt stylistic discontinuities caused by heterogeneous source footage, including
changes in color grading, film grain, and overall production quality.

\begin{table}[t]
\centering
\caption{\textbf{Quantitative evaluation on VEdit-Bench} across five dimensions:
Narrative Logic (NL), Factual Accuracy (FA), Semantic Alignment (SA), Compositional Quality (CQ), Stylistic
Cohesion (SC). Best scores among automated methods in \textbf{bold}; human editors are a reference.}
\label{tab:main_results}
\setlength{\tabcolsep}{4.5pt}
\renewcommand{\arraystretch}{1.0}
\small
\begin{tabular}{@{}lcccccc@{}}
\toprule
\textbf{Method} & \textbf{NL} & \textbf{FA} & \textbf{SA} & \textbf{CQ} & \textbf{SC} & \textbf{Overall} \\
\midrule
\quad Seedance 2.0        & 2.93 & 2.44 & 3.21 & 3.62 & 3.34 & 3.11 \\
\quad Kling 3.0           & 2.71 & 2.23 & 2.98 & 3.41 & 3.12 & 2.89 \\
\quad Veo 3.1             & 2.52 & 2.08 & 2.74 & 3.18 & 2.91 & 2.69 \\
\quad Sora 2              & 2.34 & 1.93 & 2.51 & 2.97 & 2.68 & 2.49 \\
\midrule
\quad GPT-5.4             & 3.64 & 3.82 & 3.23 & 3.11 & 3.34 & 3.43 \\
\quad Gemini 3.1 Pro      & 3.51 & 3.74 & 3.14 & 3.02 & 3.22 & 3.33 \\
\midrule
\quad MoneyPrinterTurbo   & 2.81 & 2.52 & 2.63 & 2.71 & 2.62 & 2.66 \\
\quad AutoClip            & 2.93 & 2.61 & 2.84 & 2.92 & 2.74 & 2.81 \\
\quad VideoAgent          & 3.62 & 3.34 & 3.51 & 3.43 & 3.41 & 3.46 \\
\midrule
\rowcolor{gray!12}
\quad \textbf{Auto-Director (ours)} & \textbf{4.21} & \textbf{4.34} & \textbf{4.23} & \textbf{4.02} & \textbf{4.13} & \textbf{4.19} \\
\midrule
\quad \textit{Human Editors} & \textit{4.58} & \textit{4.71} & \textit{4.52} & \textit{4.83} & \textit{4.44} & \textit{4.62} \\
\bottomrule
\end{tabular}
\end{table}

\section{Experiments}
\label{sec:experiments}

\subsection{Baselines}
\label{sec:setup}

We compare Auto-Director against four categories of methods:
\begin{itemize}[leftmargin=1em,topsep=0pt]
\setlength{\itemsep}{0pt}
 \item \textbf{Video Generation Models.} As these cannot produce long videos in one pass, we use an LLM to
 decompose the topic into per-scene captions, invoke each model per scene, and concatenate. We evaluate
 Seedance 2.0, Sora 2, Veo 3.1, and Kling 3.0.\footnote{\url{https://seed.bytedance.com/en/seedance2_0},
 \url{https://developers.openai.com/api/docs/models/sora-2},
 \url{https://gemini.google/overview/video-generation/}, \url{https://klingai.com/cn/}}
 \item \textbf{Multimodal Large Language Models.} Given the same materials and prompt, each must comprehend the
    footage, predict segment timestamps, narrate, and arrange the result. We evaluate GPT-5.4 and Gemini 3.1
    Pro.\footnote{\url{https://developers.openai.com/api/docs/models/gpt-5.4},
 \url{https://deepmind.google/models/gemini/pro/}}
 \item \textbf{Agentic Editing Frameworks.} The most direct comparison: MoneyPrinterTurbo, AutoClip, and
 VideoAgent~\cite{zhou2026videoagent}.\footnote{\url{https://github.com/harry0703/MoneyPrinterTurbo},
 \url{https://github.com/zhouxiaoka/autoclip}, \url{https://github.com/HKUDS/VideoAgent}}
 \item \textbf{Human Editors.} As a human reference, 10 professional editors with $\geq$5 years of
 post-production experience each produced a video independently from the same repository and prompt,
 without time constraints.
\end{itemize}

\subsection{Implementation Details}
\label{sec:implementation}

All agents use Gemini 3.1 Pro as the backbone model. Planning-stage retrieval uses the Google Search and
Wikipedia APIs. Scripts follow a fixed JSON schema with a predefined scene template, with at most $N=10$
scenes per video. Shot boundary detection uses PySceneDetect, and each shot is tagged with a unique Shot ID
and timestamp superimposed on keyframes sampled at 1 FPS.

First-stage retrieval uses Qwen3-Embedding~\cite{zhang2025qwen3} with $K=5$ candidates. The re-ranking weights
are $\lambda=(0.5,0.3,0.2)$ to prioritize semantic fidelity over atmospheric and temporal dimensions. ROI
stabilization uses a Kalman filter with a 5-frame smoothing window. The Critic uses a maximum iteration budget
of $\tau=3$, with thresholds $\tau_{\text{sem}}=\tau_{\text{aes}}=4$ on the 5-point scale.


\subsection{Main Results}
\label{sec:main_results}

\begin{figure}[t]
 \centering
 \includegraphics[width=0.86\linewidth]{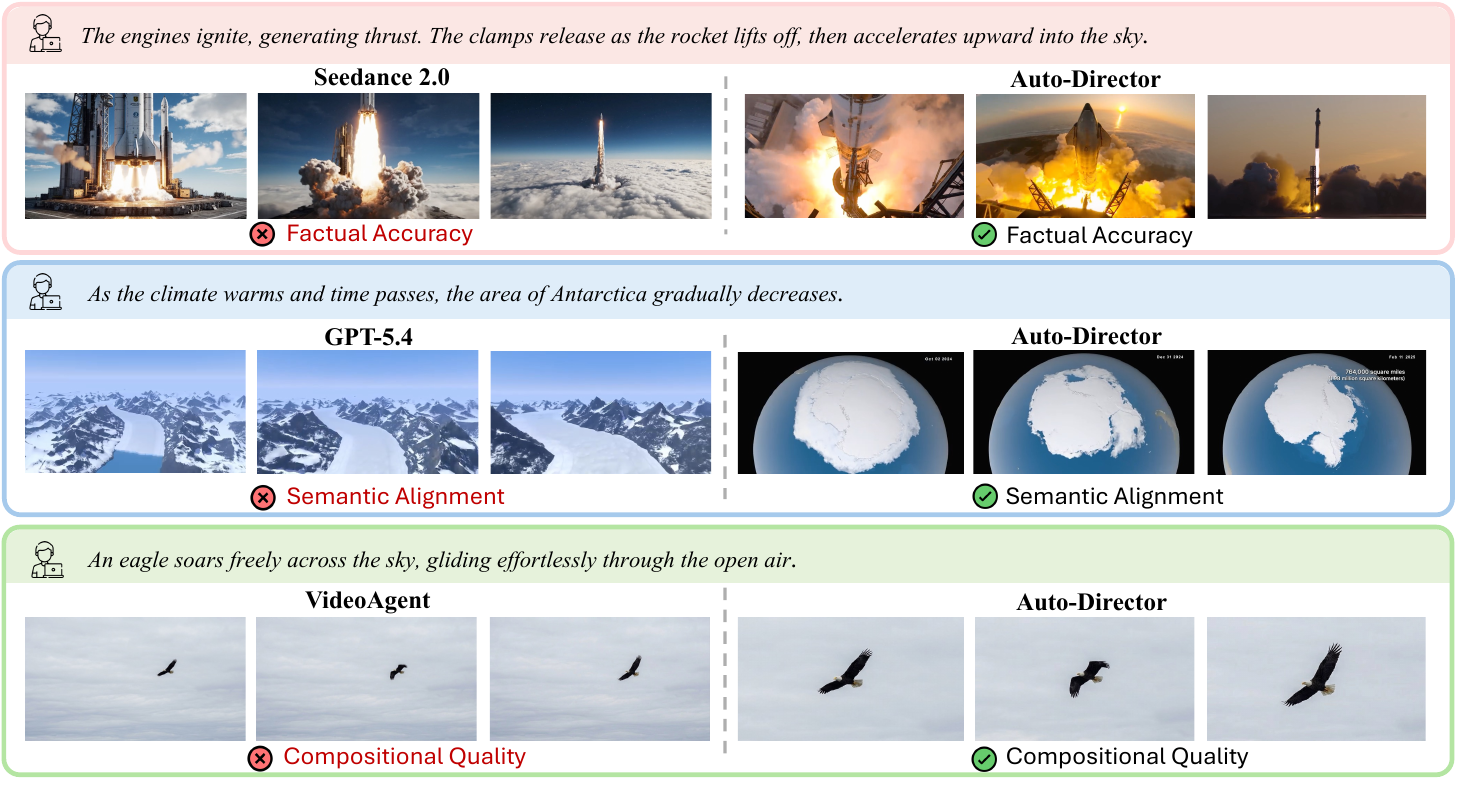}
 \caption{Qualitative comparison across three cases (factual accuracy, semantic alignment, compositional
    quality).}
 \label{fig:case-study}
\end{figure}

Table~\ref{tab:main_results} reports the comparison on VEdit-Bench. Auto-Director achieves the best performance across all five dimensions among the evaluated automated methods on all five dimensions and overall, and narrows part of the gap to professional human editors.

\begin{table}[t]
\centering
\caption{Ablation study on VEdit-Bench across the five evaluation dimensions.}
\label{tab:ablation}
\small
\setlength{\tabcolsep}{5.5pt}
\begin{tabular}{lccccc}
\toprule
\textbf{Variant} & \textbf{NL} & \textbf{FA} & \textbf{SA} & \textbf{CQ} & \textbf{SC} \\
\midrule
\textbf{Auto-Director}
& \textbf{4.21} & \textbf{4.34} & \textbf{4.23} & \textbf{4.02} & \textbf{4.13} \\
\midrule
\textit{w/o} Fact Retrieval
& 4.05 & 3.21 & 4.11 & 3.96 & 4.02 \\
\textit{w/o} Visual Tagging
& 4.03 & 4.19 & 3.54 & 3.82 & 3.71 \\
\textit{w/o} Critic Agent
& 3.91 & 3.82 & 3.85 & 3.71 & 3.64 \\
\bottomrule
\end{tabular}
\end{table}

\paragraph{\textbf{Comparison with Video Generation Models.}} These record the lowest Factual Accuracy and Narrative Logic: each per-scene clip is plausible in isolation, but concatenation provides no mechanism for cross-scene consistency, and they synthesize from parametric knowledge without verifiable evidence. The Planning Agent's Retrieval-Augmented mechanism directly addresses
this gap: by grounding every scene in externally retrieved and
cross-verified facts before any footage is assembled, Auto-Director
maintains factual accuracy that pure generation cannot replicate.

\paragraph{\textbf{Comparison with Multimodal Large Language Models.}}
Providing the same footage repository to strong MLLMs improves Factual Accuracy and Narrative Logic, with
GPT-5.4 performing best among them. However, they remain weaker on Semantic Alignment and Compositional
Quality, likely because long raw-video contexts lead to attention diffusion and coarse temporal localization.
The Material Agent instead converts footage into a compact text-indexed library $\mathcal{D}$.

\paragraph{\textbf{Comparison with Agentic Editing Frameworks.}}
VideoAgent performs best among the agentic baselines, owing to its intent decomposition and graph-based orchestration, yet still trails Auto-Director on all dimensions; MoneyPrinterTurbo and AutoClip, relying on keyword-based retrieval and single-source transcript analysis, show larger gaps in Factual Accuracy and Semantic Alignment. A key difference is the Critic's Reflect-and-Revise loop, absent in all three: instead of a one-shot output, the Critic dispatches scene-level feedback upstream, enabling targeted correction without re-running the pipeline.

\paragraph{\textbf{Gap to Human Editor.}}
Professional editors score highest on every dimension, reflecting fine-grained aesthetic judgment that is hard to automate. The gap is smallest on Semantic Alignment and largest on Compositional Quality, so sub-frame aesthetic refinement remains the clearest direction for future work; we claim no parity with human editors.

\paragraph{\textbf{Visualization Results}}
Figure~\ref{fig:case-study} presents qualitative comparisons. Generation models such as Seedance 2.0 produce
factually incorrect footage (Case 1, missing hold-down clamps), GPT-5.4-based retrieval returns plausible but
misaligned results for the Antarctica prompt (Case 2), and VideoAgent leaves the subject off-center (Case 3);
Auto-Director instead retrieves accurate physical detail, satellite views conveying the intended temporal
semantics, and composition-aware subject-centering. These are selected examples, not a quantitative summary.

\subsection{Ablation Study}
Table~\ref{tab:ablation} removes three core components individually.
Removing Fact Retrieval causes the sharpest drop in Factual Accuracy
(4.34 $\rightarrow$ 3.21) while other dimensions remain largely intact,
supporting Retrieval-Augmented Planning as the primary driver of factual grounding.
Removing Visual Tagging most severely degrades Semantic Alignment
(4.23 $\rightarrow$ 3.54), indicating that explicit shot-level annotation
matters for temporal localization.
Removing the Critic Agent produces the most uniform degradation,
consistent with the Reflect-and-Revise loop acting as a system-wide safeguard.

\subsection{User Study}
\label{sec:user_study}
\begin{table}[t]
\centering
\caption{Correlation between MLLM and human evaluation scores.}
\label{tab:user_study}
\small
\setlength{\tabcolsep}{10pt}
\begin{tabular}{lccc}
\toprule
\textbf{Dim.} & \textbf{$r$} & \textbf{$\rho$} & \textbf{MAE} \\
\midrule
NL & 0.73 & 0.70 & 0.31 \\
FA & 0.78 & 0.75 & 0.28 \\
SA & 0.75 & 0.72 & 0.30 \\
CQ & 0.68 & 0.65 & 0.38 \\
SC & 0.71 & 0.68 & 0.35 \\
\midrule
\textbf{All} & \textbf{0.73} & \textbf{0.70} & \textbf{0.32} \\
\bottomrule
\end{tabular}
\end{table}

To verify that our MLLM-based protocol produces scores consistent with human judgment, we randomly sample 20
videos and recruit 5 annotators to score each along the same five dimensions using a standardized rubric;
inter-annotator agreement reaches Krippendorff's $\alpha = 0.70$. Over all 100 scored items (20 videos
$\times$ 5 dimensions), Table~\ref{tab:user_study} reports Pearson $r = 0.73$ and Spearman
$\rho = 0.70$, demonstrating substantial alignment between automated and human evaluation.

\section{Conclusion}
\label{sec:conclusion}
We presented Auto-Director, a four-agent framework for open-world
video editing, and VEdit-Bench for evaluating the complete workflow.
Collaborative planning, material management, editing, and iterative
refinement improve performance over existing baselines across all
five evaluation dimensions.

\section*{Acknowledgments}
This work was supported in part by the Chongqing Natural Science
Foundation (CSTB2024NSCQ-MSX1026), in part by the Fundamental
Research Funds for the Central Universities
(No. SWU-KT\_XJ2025002001), in part by the National Key R\&D
Program of China (No. 2024YFC3808901), and in part by the
NSFC (62102328).

\bibliographystyle{splncs04}
\bibliography{new_bib}

\end{document}